\documentclass[letterpaper, 10 pt, journal, twoside]{IEEEtran}
\usepackage[T1]{fontenc}
\usepackage{cite}
\usepackage{amssymb,amsfonts}
\usepackage{blindtext}
\makeatletter
\let\NAT@parse\undefined
\makeatother
\usepackage{hyperref}
\usepackage{algorithmic}
\usepackage{graphicx}
\usepackage{textcomp}
\usepackage{mathtools}
\usepackage{changes}
\usepackage{subcaption}
\usepackage{algorithm}
\usepackage{xcolor}
\usepackage{color, soul}
\usepackage{amsmath}
\usepackage{booktabs}
\usepackage{multirow}
\usepackage{xstring}
\usepackage{hhline}
\usepackage{kotex}
\usepackage{siunitx}
\usepackage{comment}
\usepackage{physics}
\usepackage{tikz}
\usepackage{cleveref}
\usepackage{mathtools, nccmath}
\usepackage{cancel}

\usepackage{enumitem}

\usepackage{amssymb}
\usepackage{pifont}
\usepackage{array}
\usepackage{colortbl}

\crefformat{section}{\S#2#1#3} 
\crefformat{subsection}{\S#2#1#3}
\crefformat{subsubsection}{\S#2#1#3}

\floatname{algorithm}{Procedure}

\newcommand{\argmin}{\mathop{\mathrm{argmin}}}  
\newcommand{\argmax}{\mathop{\mathrm{argmax}}}

\definecolor{rvc}{RGB}{0, 0, 255}
\definecolor{suyong}{RGB}{0, 255, 0}
\definecolor{comment}{RGB}{0, 0, 0}
\definecolor{cv2}{RGB}{0, 0, 0}
\definecolor{qw}{RGB}{0, 0, 0}
\definecolor{qwr}{RGB}{0, 0, 0}
\definecolor{qwe}{RGB}{0, 0, 0}

\sethlcolor{lightgray}

\makeatletter
\DeclareRobustCommand{\iscircle}{\mathord{\mathpalette\is@circle\relax}}
\newcommand\is@circle[2]{%
  \begingroup
  \sbox\z@{\raisebox{\depth}{$\m@th#1\bigcirc$}}%
  \sbox\tw@{$#1\square$}%
  \resizebox{!}{\ht\tw@}{\usebox{\z@}}%
  \endgroup
}
\makeatother

\IEEEoverridecommandlockouts                              

\usepackage{caption}
\newcommand{\rom}[1]{\uppercase\expandafter{\romannumeral #1\relax}}

\title{\LARGE \bf
CLOi-Mapper: Consistent, Lightweight, Robust, and Incremental Mapper With Embedded Systems for Commercial Robot Services}

\author{DongKi Noh$^{1,2}$, \textit{Member, IEEE}, Hyungtae Lim$^{1}$, \textit{Member, IEEE}, Gyuho Eoh$^{3}$, \textit{Member, IEEE}, Duckyu Choi$^{1}$, Jeongsik Choi$^{2}$, Hyunjun Lim$^{1}$, SeungMin Baek$^{2}$, and Hyun Myung$^{1*}$, \textit{Senior Member, IEEE}
\vspace{-1.2cm}

\thanks{$^*$Corresponding author: Hyun Myung}
\thanks{$^{1}$DongKi Noh, Hyungtae Lim, Duckyu Choi, Hyunjun Lim, and Hyun Myung are with School of Electrical Engineering at KAIST (Korea Advanced Institute of Science and Technology), Daejeon, 34141, Republic of Korea. {\tt\footnotesize\{dongki.noh, shapelim, duckyu, tp02134, hmyung\}@kaist.ac.kr}}%
\thanks{$^{2}$DongKi Noh, Jeongsik Choi, and SeungMin Baek are with the Advanced Robotics Lab., CTO Division, LG Electronics, Seoul, 06772, Republic of Korea.
{\tt\footnotesize \{dongki.noh, jeongs.choi, seungmin2.baek\}@lge.com}}%
\thanks{$^{3}$Gyuho Eoh is with Department of Mechatronics Engineering, Tech University of Korea, Siheung-si, Gyeonggi-do, 15073, Republic of Korea.
{\tt\footnotesize gyuho.eoh@tukorea.ac.kr}}%
\thanks{The supplementary on our study are available at \url{https://github.com/Multiplanet-Robot}.}%
}
\begin{document}

\captionsetup[figure]{labelformat={default},labelsep=period,name={Fig.}}

\markboth{IEEE Robotics and Automation Letters. Final Version. }
{Noh \MakeLowercase{\textit{et al.}}: CLOi-Mapper: Consistent, Lightweight, Robust, and Incremental Mapper With Embedded Systems} 
\maketitle
\IEEEpeerreviewmaketitle

\begin{abstract}
In commercial autonomous service robots with several form factors, simultaneous localization and mapping (SLAM) is an essential technology for providing proper services such as cleaning and guidance. Such robots require SLAM algorithms suitable for specific applications and environments. Hence, several SLAM frameworks have been proposed to address various requirements in the past decade. However, we have encountered challenges in implementing recent innovative frameworks when handling service robots with low-end processors and insufficient sensor data, such as low-resolution 2D LiDAR sensors. Specifically, regarding commercial robots, consistent performance in different hardware configurations and environments is more crucial than the performance dedicated to specific sensors or environments. Therefore, we propose \textbf{a)} a multi-stage 
approach for global pose estimation in embedded systems; \textbf{b)} a graph generation method with zero constraints for synchronized sensors; and \textbf{c)} a robust and memory-efficient method for long-term pose-graph optimization. As verified in in-home and large-scale indoor environments, the proposed method yields consistent global pose estimation for services in commercial fields. Furthermore, the proposed method exhibits potential commercial viability considering the consistent performance verified via mass production and long-term ($>$~5~years) operation.
\end{abstract}

\vspace{-0.25 cm}
\section{Introduction} \label{sec:intro}
\begin{figure}[t!]
    \captionsetup{font=footnotesize}
    \centering
	\begin{subfigure}[b]{0.45\textwidth}
		\includegraphics[width=1\textwidth]{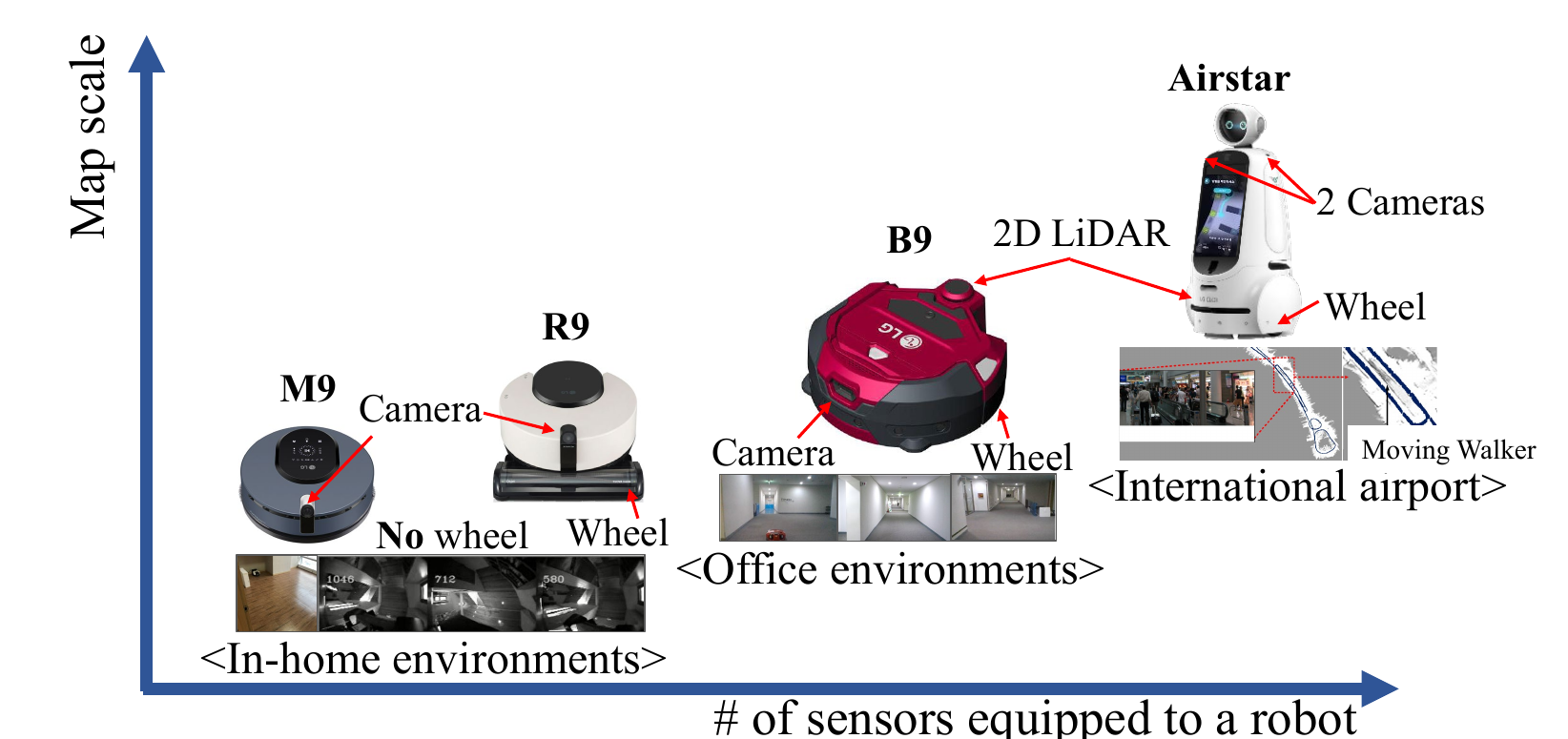}
            \vspace{-0.5cm}
            \caption{}
	\end{subfigure}

        \begin{subfigure}[b]{0.40\textwidth}
		\includegraphics[width=1\textwidth]{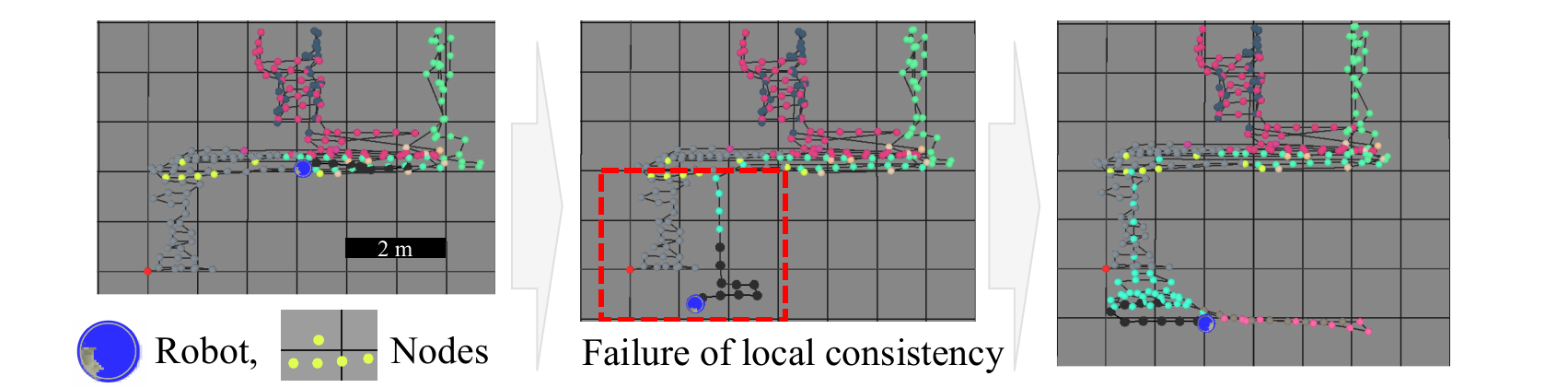}
            \vspace{-0.5cm}
            \caption{ORB-SLAM~\cite{campos2021orb} with our embedded system (R9)}
      \end{subfigure}	
 
        \begin{subfigure}[b]{0.40\textwidth}
		\includegraphics[width=1\textwidth]{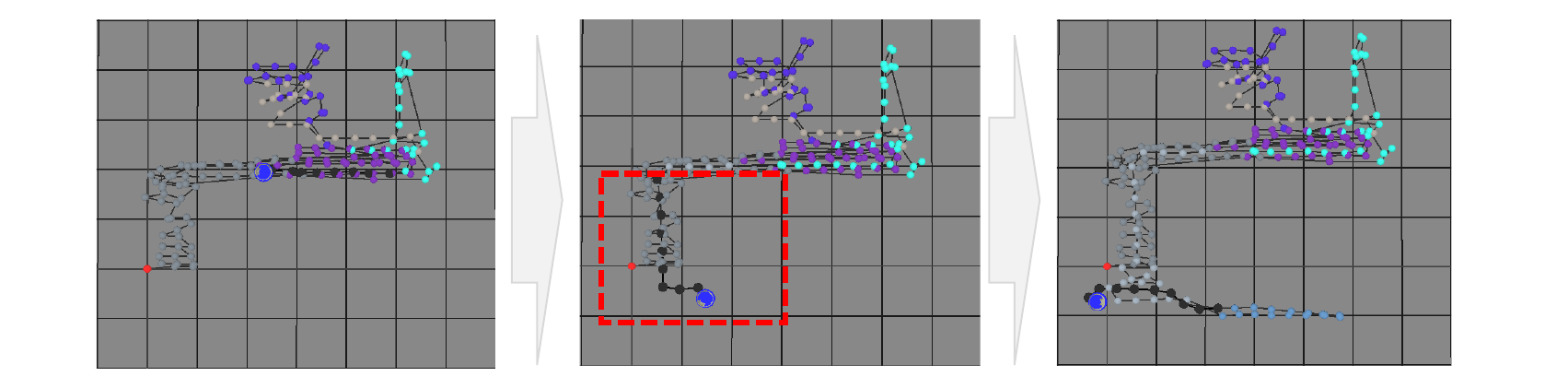}
            \vspace{-0.5cm}
            \caption{CLOi-Mapper with our embedded system (R9)}
	\end{subfigure}		
 
\caption{(a) Illustration of our SLAM applications arranged from bottom-left to top-right. Airstar is the world's first commercially available airport guidance robot utilized in an international airport. Although each robot has a different sensor configuration, computational performance, and operational condition, they all utilize the proposed framework. (b)--(c) Visualization of the global pose tracking output in an indoor space. Empirically, it was demonstrated that CLOi-Mapper exhibits less delayed pose tracking performance with the proposed embedded processor ($\leq$ 1 GFLOPS) (depicted in the red dotted rectangle), thus enabling the proposed system to enhance trajectory smoothness by mitigating duplicated trajectories and large corrections. In addition, the colors of nodes represent nodes within the same local map.}
	\label{fig:fig_CommercialRobots_in_LGE}
	\vspace{-0.6cm}
\end{figure}

\IEEEPARstart{S}{imultaneous} localization and mapping (SLAM) has long been an essential technique in robot navigation and autonomous platform fields. To apply SLAM algorithms to real-world robots, various methods with exteroceptive sensors, such as range, vision, and depth sensors, have been widely utilized~\cite{10107752,wang2023sw,bai2022faster, xu2022fast, sung2022if, song2022dynavins, qin2018vins, leutenegger2015keyframe, campos2021orb, mur2015orb, KohlbrecherMeyerStrykKlingaufFlexibleSlamSystem2011, labbe2019rtab, hess2016real}. 
Recently, there has been a growing need to apply SLAM algorithms to service robots with diverse sensor configurations, as illustrated in Fig. ~\ref{fig:fig_CommercialRobots_in_LGE}(a). Moreover, consistent real-time service has become crucial at both the mapping and localization stages. Regarding the failure of local consistency, as illustrated in Fig.~\ref{fig:fig_CommercialRobots_in_LGE}(b), robot services might be temporarily halted owing to significant position corrections. However, despite encountering limitations in implementing innovative algorithms within embedded systems, the proposed method has been successfully applied to real service robots, thus enabling consistent pose estimation and mapping, as illustrated in Fig.~\ref{fig:fig_CommercialRobots_in_LGE}(c).

Accordingly, this study introduces various relevant requirements for applying SLAM algorithms to real-world service robots and presents approaches for handling them. Considering commercial service robots, we collected several requirements from researchers, developers, and marketers in service robot communities, and summarized them into three: a) SLAM algorithms should be extensible to include complementary sensors and ensure stable services in different environments, b) locally consistent global poses and a globally consistent map should be provided in both localization-only and SLAM stages, and c) SLAM algorithms applied to service robots should exhibit stable performance even in resource-restricted systems. In our case, we have a lack of memory ($\leq$ 200 MB), low-frequency inertial measurement unit (IMU) data transfer ($\leq$ 20 Hz) due to narrow bandwidth between an application processor (AP) and a micro controller unit, and a low-cost processor ($\leq$ 1 GFLOPS).

Various methods can be employed to satisfy local consistency of pose estimation in real-time, such as the LiDAR-inertial odometry~(LIO)~\cite{wang2023sw,bai2022faster,xu2022fast,sung2022if}, visual-inertial odometry~(VIO)~\cite{qin2018vins, song2022dynavins, leutenegger2015keyframe,campos2021orb}, and sensor fusion methods~\cite{10107752,lee20232, 9561996, zhao2021super}. However, these frameworks require computational capabilities for tracking poses across multiple sequential frames (e.g. bundle adjustment (BA)), which could be challenging to implement in embedded systems.

Accordingly, this study focuses on extensible methods to fuse sensors complementarily in different combinations; unlike existing studies, this letter presents more practical methods for providing locally consistent global poses in real-time for location-based services with mobile robots. In particular, we leveraged the Kalman-filter-like tracking method based on the Bayesian framework with a minimum number of nodes, and verified that the proposed method can be applied to systems with a low-frequency IMU and embedded processor. In addition, regarding globally consistent poses, we employ an efficient graph generation method with temporal node integration and node pruning to mitigate the degradation of measurements from low-cost sensors and handle memory limitations in resource-constrained systems.

As verified in myriad in-home and large-scale indoor environments, the proposed method, named as CLOi-Mapper, an abbreviation of Consistent, Lightweight rObust incremental Mapper, has been applied to various commercialized services. In particular, to the best of our knowledge, Airstar is the first commercialized robot (since 2018) that has been operational as an airport guidance robot in the real world for more than five years. Although European researchers conducted a study on a guidance robot in the Spencer project\footnote[1]{\url{http://www.spencer.eu}} at Schiphol International Airport, Netherlands, it is yet to be commercialized. In conclusion, the framework presented herein represents the culmination of a 5-year development process, continuously improving the initial version established five years ago. The contributions of this letter are as follows: 
\begin{itemize}
\vspace{-0.05cm}
\item A novel mapping system is proposed, which comprises a simplified Bayesian framework-based method for consistent pose estimation and efficient back-end suitable for resource-constrained systems, e.g. low-end sensors and processors with insufficient memory.
\item Moreover, the zero-constraints-based graph structure is presented, which can handle different combinations of sensor types and form factors with minimal tuning.
\item In particular, the generalization capability of the proposed method has been verified via mass production.
\vspace{-0.05cm}
\end{itemize}

The remainder of this paper is organized as follows. Section~\ref{sec:survey} introduces SLAM algorithms in terms of commercialized service robots. Section~\ref{sec:Versatile Online SLAM Framework} comprehensively elucidates the proposed CLOi-Mapper. Subsequently, Sections~\ref{sec:Experiments}~and~\ref{sec:Experimental Results} present the experimental setups and results for real-world robots, respectively. Finally, a summary of our findings and conclusions are presented in Section~\ref{sec:conclusion}.

\section{Related Works}\label{sec:survey}

\subsection{SLAM Applications to Service Robots}

According to World Robotics Reports~\cite{worldrobotics}, several sensor configurations related to SLAM are utilized in service robots, e.g. cameras, RGB-D, LiDAR, and radar sensors. In particular, SLAMTek is well-known for LiDAR-based solutions employed in various service robots. Dyson and iRobot introduced a visual feature-based SLAM with an omni-view and forward-view camera. Artificial markers, such as QR codes, can be partially utilized to ensure that indoor delivery services are reliable in complex environments.

Although the algorithms applied to robots introduced in the World Robotics Reports have rarely been released with source codes, several approaches with low-cost sensors and IMUs are expected to be suitable for embedded systems. ORB-SLAM~\cite{mur2015orb} series is one of the most popular visual feature-based SLAM systems that can operate with monocular, stereo, and RGB-D cameras. Grid map-based algorithms similar to Hector-SLAM~\cite{KohlbrecherMeyerStrykKlingaufFlexibleSlamSystem2011} have been applied to various commercialized cleaning robots with 2D LiDAR sensors. Recently, RGB-D sensors and algorithms similar to RTAB-Map~\cite{labbe2019rtab} have been utilized to generate dense 3D maps of indoor environments. In addition, Google Cartographer~\cite{hess2016real} functions as a real-time SLAM system that enables robots to accurately map large-scale environments with a 2D LiDAR sensor and an affordable embedded system.  

Until recently, low-cost hardware remains prevalent in mass-produced consumer robotic products. In this context, this study focuses on algorithms capable of managing diverse SLAM applications with combinations of low-cost hardwares and sensors.

\vspace{-0.3cm}
\subsection{Studies on Sensor Fusion and Real-Time Operation}
\vspace{-0.1cm}

Recently, Shan~\textit{et al.}~\cite{9561996} proposed efficient sensor fusion with multiple sensors using the factor-graph. Moreover, Zhao~\textit{et al.}~\cite{zhao2021super} presented an easier and more flexible method to fuse multiple sensors robustly in a mini computer with sub-factor graphs. However, these algorithms were not applied to commercialized service robots with an embedded processor, but to a platform with an Intel NUC onboard computer; hence, this approach may be unsuitable for low-cost embedded systems. Moreover, even service robot systems with hardware described in~\cite{zhao2021super} inherently exhibit temporal latency characteristics in low-priority tasks because they simultaneously handle several devices and tasks, including SLAM. Consequently, recent studies~\cite{sola2022wolf,cramariuc2022maplab} indicate that synchronization can be a challenging problem in real robotic systems. 
\vspace{-0.3cm}
\section{CLOi-Mapper}\label{sec:Versatile Online SLAM Framework}

\subsection{System Overview}
\begin{figure}[t!]
    \vspace{0.25cm}
    \captionsetup{font=footnotesize}
    \centering
	\begin{subfigure}[b]{0.45\textwidth}
		\includegraphics[width=1\textwidth]{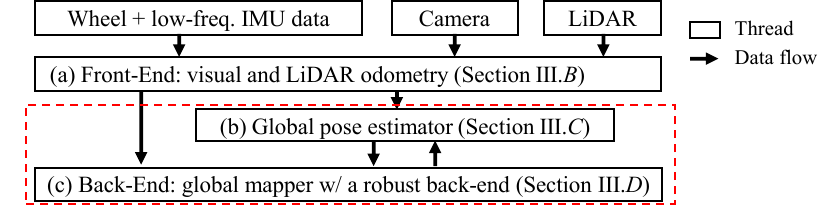}
	\end{subfigure}
	\caption{Block diagram presenting an overview of CLOi-Mapper. The dashed box represents the novel and modified blocks proposed by this study. The functionalities of each block are described as follows: (a) Front-end of our framework with visual and LiDAR odometry, (b) semi-real-time global pose tracker with a simplified graph, (c) global mapper with pose-graph optimization, including a temporal node and the novel pruning method.}
	\label{fig:fig_Pipeline}
	\vspace{-0.6cm}
\end{figure}

The proposed method addresses locally and globally consistent pose estimation of ground mobile robots with resource-restricted systems targeted to a commercial level. Based on the aforementioned requirements, the proposed CLOi-Mapper primarily comprises three parts, as illustrated in Fig.~\ref{fig:fig_Pipeline}. 

First, the proposed method begins with extensible graph generation to handle various sensor configurations without modifying the framework. At this stage, frame nodes related to visual and LiDAR sensors are generated by the method introduced in Section~\rom{3}.\textit{B}. In particular, we propose a \textit{zero-constraint} to handle synchronized sensors (see Section~\rom{3}.\textit{B}.3). 

Second, given these frame nodes, this study aims to estimate locally and globally consistent global poses. In general, batch optimization applied to a graph-based SLAM can provide accurate state estimation results. Nevertheless, it requires high computational complexity, which potentially leads to system latency. Hence, we propose a  Kalman-filter-like tracking method with an anchor node to yield locally consistent poses for real-time robot operation, even in the worst-case scenario (see Section~\rom{3}.\textit{C}). 

Finally, to improve memory efficiency and simplify graph optimization for real-time operation, this study involves pruning and merging the graph comprising locally consistent poses. Specifically, our proposed method uses both geometric and information weights derived from a grid map to maintain a geometrically uniform distribution of nodes (see Section~\rom{3}.\textit{D}.2).

We adopted these modules and proposed a multi-stage method for globally consistent pose estimation. These modules are explained in the following subsections. 

\vspace{-0.3cm}
\subsection{Front-End in CLOi-Mapper}\label{sec:VisualOdometry}
\subsubsection{Visual Odometry for Visual Frames}\label{sec:VisualOdometry}

Visual odometry aims to estimate relative poses between keyframes using the geometrical relationship between 2D points ($\mathbf{p}_{p}$) in pixel coordinates
~and 3D ($\mathbf{p}_{w}$) points in world coordinates. Keyframes can be selected among given frames with several rules, such as sufficient 3D/2D points for feature matching, redundancy check of a frame based on the $\chi^2$-test, and distance traveled.
We utilized a general projective camera model~\cite{kannala2006generic} to determine the relationship between 2D and 3D points. The 3D points in the $(k-1)$-th frame can be reprojected into the 2D points in the $k$-th frame by $\mathbf{p}_{p,k} = {\pi}(\mathbf{\mathcal{M}}_{k}, \mathbf{p}_{w,k-1}),$ where $\mathbf{p}_{p,k}$, $k$, $\mathbf{\mathcal{M}}_{k}$, and ${\pi}(\cdot)$ denote a reprojected point on the pixel coordinate in the $k$-th frame, a keyframe index, $k$-th projection matrix, and the reprojection function of 3D point, respectively. By employing~$\mathbf{p}_{p,k}$ and the reprojection error estimated via sparse bundle adjustment (SBA)~\cite{lourakis2009sba}, the odometry between sequential visual frames in a sliding window can be estimated. 

Empirically, the number of reliable features may be limited owing to the utilization of a cost-effective camera in our system. To address this limitation and ensure consistent visual odometry (VO) performance in scenarios with limited features, we suggest three methods: a) we projected 3D points from the $k$-th and $(k+1)$-th keyframes onto the $(k-1)$-th keyframe as 2D points. This projection allows the $(k-1)$-th keyframe to incorporate supplementary measurements (2D/3D features) from the $k$-th and $(k+1)$-th keyframes, thereby enhancing the accuracy of 3D points and poses, b) we merged duplicated 3D points between consecutive keyframes to enhance accuracy by averaging the corresponding points, and c) to eliminate less informative features and improve the accuracy of features, we implemented a filter to discard 3D points observed in only a few keyframes, typically less than four in indoor scenarios, based on empirical observations.

\subsubsection{LiDAR Odometry for LiDAR Frames and 2D Mapping}\label{sec:LiDAROdometry}
We leveraged iterative closest point (ICP)~\cite{besl1992method} to estimate odometry between sequential LiDAR frames and generate loop constraints. LiDAR frames are generated independently at every timestamp. However, not all LiDAR frames can be utilized owing to computational limitations in real robotic systems. Therefore, LiDAR keyframes are selected based on two conditions: completing the optimization processes of VO and LO, and being farther than a predefined distance from the center of an adjacent LiDAR frame in the global coordinate.

\begin{figure}[t!]
\vspace{0.25cm}
    \captionsetup{font=footnotesize}
    \centering
	\begin{subfigure}[b]{0.48\textwidth}
		\includegraphics[width=1\textwidth]{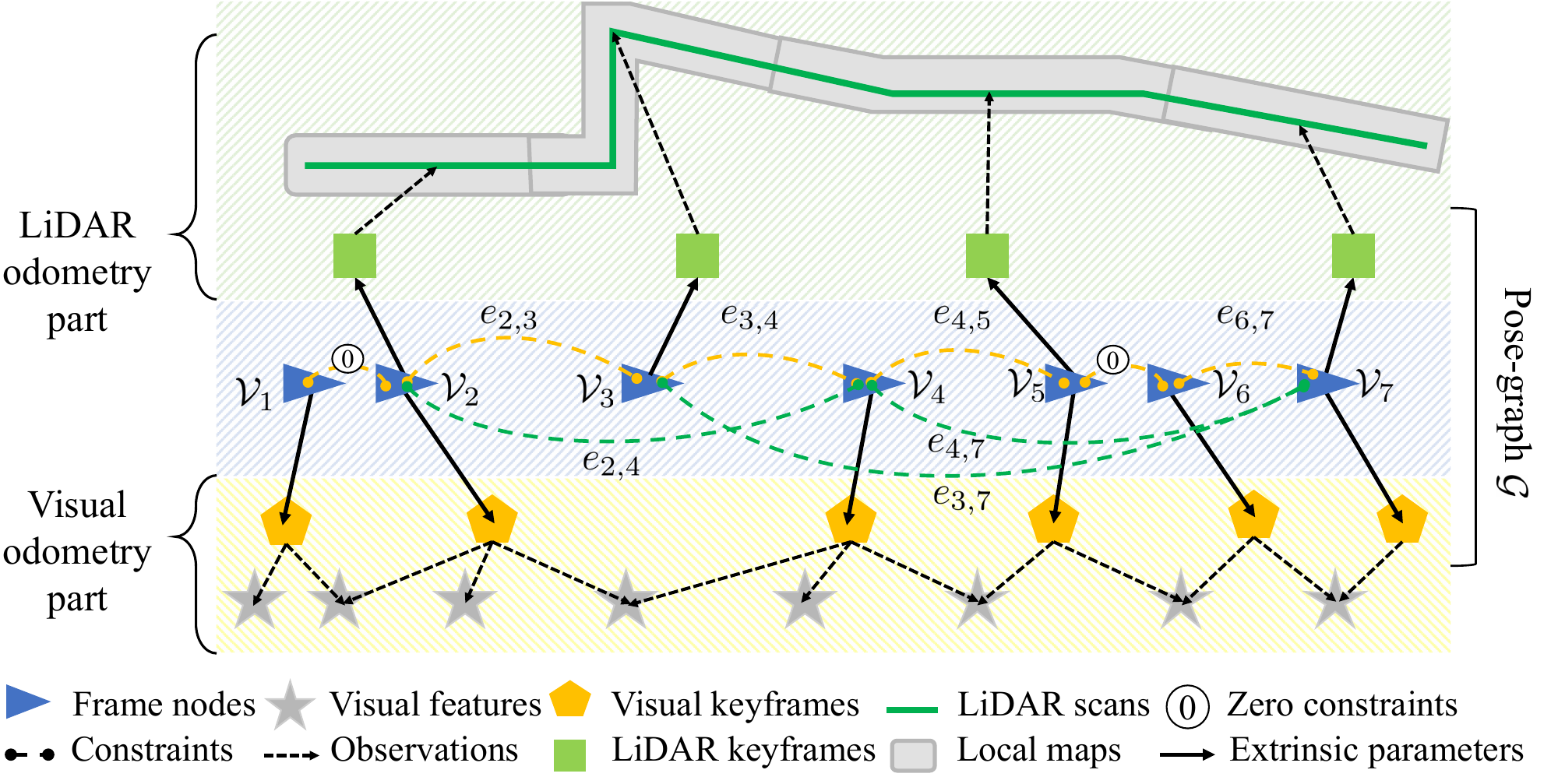}
	\end{subfigure}
	\caption{Overview of a structurally extensible pose graph with cameras and a 2D LiDAR sensor. Visual and LiDAR odometry parts contain visual features and local maps, respectively, and the two parts are interconnected via the frame nodes set $\mathcal{V}$. Frame nodes comprise a combination of wheel odometry with IMU, visual odometry, and LiDAR odometry.} 
	\label{fig:fig_PoseGraph}
	\vspace{-0.5cm}
\end{figure}
\subsubsection{Extensible Graph Generation for Handling Various Sensor Configurations}
The pose-graph is defined as $\mathcal{G}$~=~$(\mathcal{V},\mathcal{E})$, where $\mathcal{V}=\{\mathcal{V}_1,\mathcal{V}_2,\dots,\mathcal{V}_N \}$ is a set of frame nodes $\mathcal{V}_{a}~(1 \leq a \leq N)$ and $\mathcal{E} \subset \mathcal{V} \cross \mathcal{V}$ is a set of ordered pairs $\{ e_{a, b}\}_{a \neq b}$ between nodes $\mathcal{V}_{a}$ and $\mathcal{V}_{b}~(1 \leq b \leq N)$, where $N$ is the total number of frame nodes. Using visual and LiDAR frames, the proposed method indirectly connects nodes generated by distinct sensors by linking through the intermediary frame nodes set denoted as $\mathcal{V}$, as illustrated in Fig.~\ref{fig:fig_PoseGraph}. 

\subsubsection{Edge Generation in the Case of Synchronized Sensors}

\begin{figure}[t!]
\vspace{0.25cm}
    \captionsetup{font=footnotesize}
    \centering
	\begin{subfigure}[b]{0.40\textwidth}
		\includegraphics[width=1\textwidth]{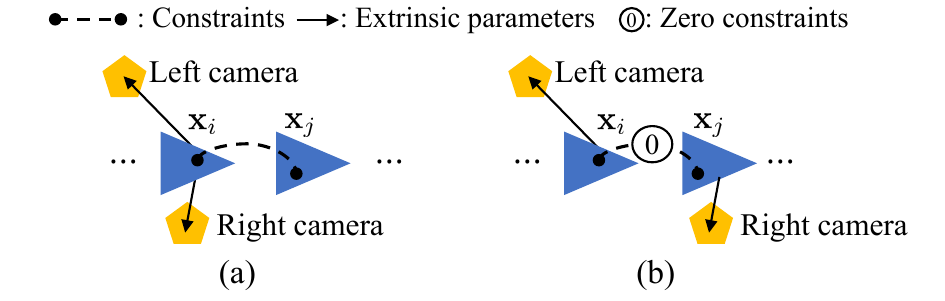}
	\end{subfigure}
	\vspace{-0.2cm}
	\caption{Illustration of the proposed \textit{zero-constraint} concept. (a) Regarding the robot with synchronized cameras, various studies employ image merging or fusion of odometry estimated by each camera. (b) In contrast, we propose a zero-constraint, setting it as a relative pose to enable the synchronization effect without altering the framework.}
	\label{fig:fig_zeroconstraints}
	\vspace{-0.6cm}
\end{figure}

We propose a method for handling various sensor configurations without altering the framework. Let us assume that cameras are already synchronized; then, we should handle the sensor data of cameras simultaneously. Accordingly, data stacking methods, such as image stitching, should be employed in such cases, which leads to the framework's modification. This could be a limitation when managing various robots under a unified framework for efficiency. To address this limitation, we assume that synchronized sensor data are obtained sequentially, as illustrated in Fig.~\ref{fig:fig_zeroconstraints}. Hence, we set a relative pose between them to the identity matrix, which we called the \textit{zero-constraint}. To employ this approach with numerical stability, this study adopts a strategy where we set a covariance value close to zero and employ an information matrix based on a heuristic approach, which is associated with the determinant-based method described in~\cite{rodriguez2018importance} as follows:\useshortskip 
\begin{equation}
I(\mathbf{x}_i;\mathbf{x}_j) \approx \begin{cases}\|\mathbf{\Sigma}_{i,j}^{-1}\| & \text { if } \mathbf{\Sigma}_{i,j}^{-1}~\text{exists} \\ |1/\operatorname{det}(\mathbf{\Sigma}_{i,j})| & \text { if }  \mathbf{\Sigma}_{i,j}^{-1} \rightarrow \infty \end{cases},
\label{eqn:mutual_information}
\vspace{-0.1cm}
\end{equation}
where $I$, $\boldsymbol{\Sigma}_{i,j}$, and $\operatorname{det}(\cdot)$ denote mutual information for generating Chow-Liu trees~\cite{chow1968approximating,kretzschmar2012information} in our back-end, covariance between $i$-th and $j$-th nodes, and a determinant, respectively. 
\vspace{-0.5cm}
\subsection{Semi-Real-Time Global Pose Estimation in CLOi-Mapper}\label{sec:GPE}
\subsubsection{Bayesian-Framework-Based Simplified Global Pose Estimation}
Pose graph optimization (PGO) for global pose estimation could be computationally intensive in embedded systems. Hence, PGO is executed with the lowest priority in practical robotic systems equipped with an embedded processor to mitigate total system delay. Consequently, the execution of PGO for global pose estimation is inevitably delayed, leading to pose drift.

\begin{figure}[t!]
    \vspace{0.25cm}
    \captionsetup{font=footnotesize}
    \centering
	\begin{subfigure}[b]{0.5\textwidth}
		\includegraphics[width=1\textwidth]{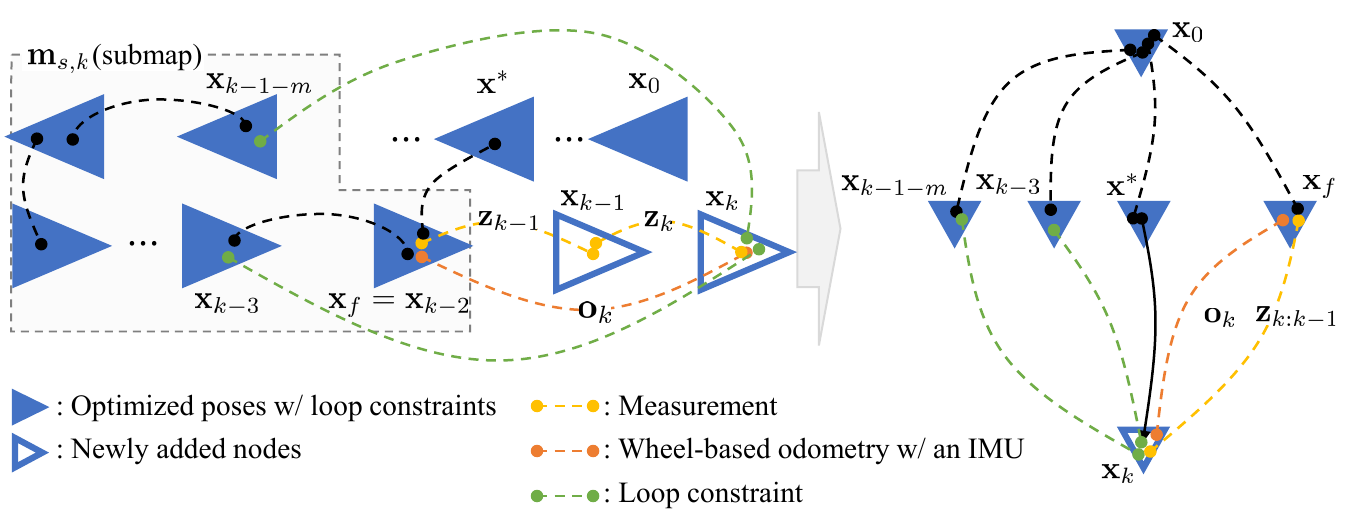}
	\end{subfigure}
	\vspace{-0.2cm}
 \vspace{-0.5cm}
	\caption{Illustration of simplified pose graph generation between the reference pose ($\mathbf{x}_{0}$) and current pose ($\mathbf{x}_{k}$) in global coordinates. (L-R) By utilizing the metric embedding method~\cite{lim2011online}, we transform the original pose graph into the simplified pose graph with previously optimized nodes, odometry $\mathbf{o}_{k}$, measurements $\mathbf{z}_{k-1:k}$, and loop constraints in the SLAM submap ($\mathbf{m}_{s,k}$).}
	\label{fig:fig_GPT_graph}
	\vspace{-0.7cm}
\end{figure}

Nevertheless, in order to track a desired trajectory consistently in time-delayed systems, this letter proposes a simplified Bayesian approach for optimization. Let the conditional probability density function (PDF) of the latest global pose ($\mathbf{x}_{k} \in\mathbb{R}^3$) be $p(\mathbf{x}_{k}| \mathbf{x}_{0:k-1},\mathbf{m},\mathbf{z}_{1:k})$, where $\mathbf{x}_{0:k-1}$, $\mathbf{z}_{1:k}$, and $\mathbf{m}$ represent the set of poses, the set of measurements, and the entire SLAM map, respectively. We employ a previously optimized node as an anchor node, denoted as $\mathbf{x}^{\ast}=\argmin_{\mathbf{x}} \text{SMD}(\mathbf{x}_{f},\mathbf{x})$, where SMD and $\mathbf{x}_{f}$ represent the squared Mahalanobis distance and last pose optimized by PGO, respectively, as illustrated in Fig.~\ref{fig:fig_GPT_graph}. The anchor node enables the new robot pose to maintain consistency with the previously optimized robot poses. By using a Bayesian approach and leveraging the local adjustment module~\cite{lim2011online}, $p(\mathbf{x}_{k}| \mathbf{x}_{0:k-1},\mathbf{m},\mathbf{z}_{1:k})$ can be approximated as 
\begin{equation}
  \begin{aligned}
    p(\mathbf{x}_{k}| \mathbf{x}_{k-1},\mathbf{x}_f,\mathbf{x}^{\ast},\mathbf{x}_{0},\mathbf{m}_{s,k},\mathbf{z}_{1:k}),
  \end{aligned}
  \label{eqn:pdfofacurrentPose}
\end{equation}
where $\mathbf{x}_{0}$, $\mathbf{x}_{k}$, and $\mathbf{m}_{s,k}$ denote a reference pose, the pose of the $k$-th frame node $\mathcal{V}_{k}$, and a SLAM submap that is a covisibility map around $\mathbf{x}_{k}$, respectively. Compared with LIO and VIO~\cite{10107752,wang2023sw,bai2022faster, xu2022fast, bloesch2017iterated}, the proposed method exclusively depends on the minimum number of nodes such as $\mathbf{x}_{0}$ and $\mathbf{x}_{k}$, resulting in real-time pose estimation. As illustrated in~Fig.~\ref{fig:fig_GPT_graph}, the conditional PDF in~(\ref{eqn:pdfofacurrentPose}) can be represented with odometry, measurements, and loop constraints within a submap $\mathbf{m}_{s,k}$. By using this representation, (\ref{eqn:pdfofacurrentPose}) is factorized as follows:  

\begin{equation}
  \begin{aligned}     \underbrace{p(\mathbf{x}_{k}|\mathbf{\bar{x}}_{k},\mathbf{z}_{k-1:k})}_{\text{Measurement}}\underbrace{p(\mathbf{\bar{x}}_{k}|\mathbf{x}_f,\mathbf{o}_{k})}_{\text{Motion model}}p(\mathbf{x}_{0})\boldsymbol{\alpha}_k\prod_{q}\rho(\boldsymbol{\beta}_{q,k})\\ \text{s.t.}~\boldsymbol{\alpha}_k=\exp_{t}(\mathbf{\bar{x}}_{k},\mathbf{x}^{\ast})~\text{and}~\boldsymbol{\beta}_{q,k}=\exp_{t}(\mathbf{\bar{x}}_{k},\mathbf{x}_{q}),
  \end{aligned}
  \label{eqn:manipulatedpdfofacurrentPose}
\end{equation}
where $\mathbf{\bar{x}}_{k}$, $\mathbf{o}_{k}$, and ${\rho}$ denote a predicted pose, wheel-based odometry fused by IMU's angular rate, and a \textit{Huber} norm, respectively; pairs ($q,k$) belong to a set $\gamma$ of visual and LiDAR loop constraints. In addition, $\exp_{t}(\mathbf{x}_m,\mathbf{x}_n)$ is the Gaussian uncertainty model between a pose $\mathbf{x}_m$ and $\mathbf{x}_n$, defined as $\frac{1}{\eta}\exp{-\text{SMD}(\mathbf{x}_m,\mathbf{x}_n)/2}$, where $\eta$ denotes a normalization factor. Finally, \textcolor{qwr}(\ref{eqn:pdfofacurrentPose}) can be simplified to an optimization problem between the current node ($\mathbf{x}_{k}$) and the reference node ($\mathbf{x}_{0}$), which is graphically illustrated in Fig.~\ref{fig:fig_GPT_graph}. By manipulating \textcolor{qwr}(\ref{eqn:manipulatedpdfofacurrentPose}) with a negative logarithm, the solution to \textcolor{qwr}(\ref{eqn:manipulatedpdfofacurrentPose}) can be obtained by minimizing the sum of residuals as follows:
\begin{equation}
\begin{aligned}
  \mathbf{\hat{x}}_{k}=\argmin_{\mathbf{x}_{k}} \Big{\{} \boldsymbol{\zeta}_k + \log(\boldsymbol{\alpha}_k) + \sum_{(q,k) \in \mathbf{\gamma} } \log\big{(}{\rho}(\boldsymbol{\beta}_{q,k}) \big{)} \Big{\}}\\ \text{s.t.}~\boldsymbol{\zeta}_k = \| h(\mathbf{x}_{k},\mathbf{x}_{0})-\mathbf{z}_{k,0}\|^{2}_{\mathbf{\Sigma}_{k,0}},
  \end{aligned}
  \label{eqn:costfunctiononGPT}
\end{equation}
where ${h}(\cdot)$, $\mathbf{z}_{k,0}$, and $\mathbf{\Sigma}_{k,0}$ represent the function of odometry prediction between two sequential nodes, observations pertaining to odometry, and combined covariance matrices~\cite{smith1986representation}, respectively.

\subsubsection{Pose Prediction in the Worst Case}

In the worst case, computational limitations could delay the aforementioned global pose estimation, thereby increasing the bound of robot pose error and leading to local consistency degradation. To address this situation, we assume that the $(k+1)$-th transform matrix $\delta \mathbf{T}_{k+1}$ for correction of wheel-based odometry is very similar to the previous one $\delta \mathbf{T}_{k}$. Consequently, if the global pose estimation is not conducted within the predefined time, we apply the previous correction transform $\delta \mathbf{T}_{k}$ to adjust the odometry at the current step as follows:  

\begin{equation}
\begin{aligned}
    \bar{\mathbf{T}}_{k+1} &= \delta \mathbf{T}_{k+1} \oplus \mathbf{T}_{\text{od},k+1} \approx \delta \mathbf{T}_{k} \oplus \mathbf{T}_{\text{od},k+1},
\end{aligned}
\label{eqn:odometry_cascading}
\end{equation}
where $\bar{\mathbf{T}}$, $\mathbf{T}_{\text{od}}$, and the operator $\oplus$ denote a predicted pose transform matrix, wheel-based odometry pose transform matrix, and the composition, respectively. With the previous corrected pose $\hat{\mathbf{T}}_{k} = \delta \mathbf{T}_{k}\oplus \mathbf{T}_{\text{od},k}$ and odometry propagation $\mathbf{T}_{\text{od},k+1} = \mathbf{T}_{\text{od},k} \oplus (\Delta \mathbf{T}_{\text{od},k+1})$, we manipulate (\ref{eqn:odometry_cascading}) as follows: 
\begin{equation}
\begin{aligned}
    \delta \mathbf{T}_{k} \oplus \mathbf{T}_{\text{od},k+1} &= (\hat{\mathbf{T}}_{k} \ominus \cancel{\mathbf{T}}_{\text{od},k}) \oplus (\cancel{\mathbf{T}}_{\text{od},k} \oplus \Delta \mathbf{T}_{\text{od},k+1})\\
    &=\hat{\mathbf{T}}_{k} \oplus \Delta \mathbf{T}_{\text{od},k+1},
\end{aligned}
\label{eqn:odometry_prediction}
\end{equation}
where the operators $\ominus$ and $\Delta$ denote the inverse composition and a relative pose, respectively. 

Utilizing a predicted pose transform matrix in (\ref{eqn:odometry_prediction}) as an alternative to the corrected pose transform matrix during time-delayed situations might be a viable approach to maintain the desired trajectory without significant discontinuity.

\vspace{-0.3cm}
\subsection{Graph Pruning-Based Robust Back-End in CLOi-Mapper}\label{sec:Robust Back-end with Graph Pruning and Optimization}
To enhance memory efficiency and graph optimization robustness in real robotic systems, this letter proposes a back-end with a pruning and merging method.

\subsubsection{Graph Generation With a Temporal Node}

\begin{figure}[t!]
\vspace{0.25cm}
    \captionsetup{font=footnotesize}
    \centering
	\begin{subfigure}[b]{0.45\textwidth}
		\includegraphics[width=1\textwidth]{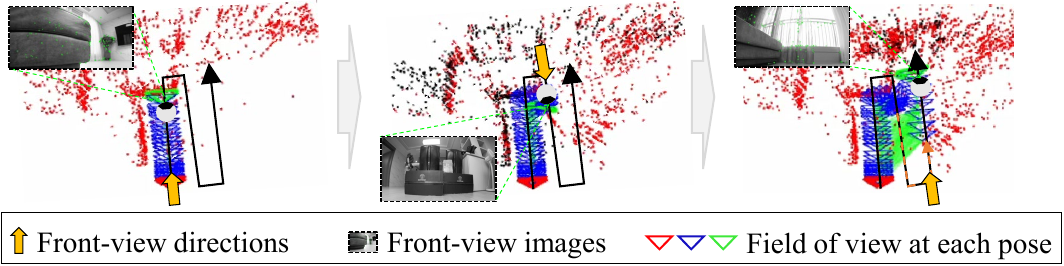}
	\end{subfigure}
	\vspace{-0.2cm}
	\caption{Illustration of an example of the pose tracking failure (a dotted orange line) of ORB-SLAM3, depicting in-home conditions featuring zigzag motion and pure rotation. Black arrows and accompanying images represent the robot's trajectory and forward views captured by its onboard camera.}
	\label{fig:fig_ORB_Tracking_fail}
	\vspace{-0.7cm}
\end{figure}

Owing to various harsh environments, pure rotation caused by zigzag motions, and low-cost sensors, reliable keyframes may not be consistently generated in practical SLAM applications. Typically, these unreliable keyframes, such as a frame with few features, are prone to skipping due to their lower measurement quality, potentially leading to pose tracking failures, as illustrated in Fig.~\ref{fig:fig_ORB_Tracking_fail}. To mitigate such failures within sequential keyframes, we propose a graph generation approach to best exploit unreliable keyframes (e.g. those with insufficient features). We refrain from immediately skipping keyframes with insufficient features. Instead, we subject them to optimization and subsequently assess their quality to determine their usefulness in the node pruning stage.

We define sets of unregistered (new) and optimized frame nodes as $\mathcal{V}_{\text{new}}=\{\mathcal{V}_{k-M+1}, \mathcal{V}_{k-M+2}, \dots, \mathcal{V}_{k}\}$ and $\mathcal{V}_{\text{opt}}=\{\mathcal{V}_{0}, \mathcal{V}_{1}, \dots, \mathcal{V}_{k-M}\}$ in the global map, respectively, where $M$ represents the window size of a pose graph. Let $\mathcal{V}_{f}= \mathcal{V}_{k-M+1}$ be the first frame of a set of unregistered frames as a temporal frame node. $\mathcal{V}_{f}$ contains a pose transform matrix $^{w}\mathbf{T}_{f}$ and covariance $\mathbf{\Sigma}_{f}$. $^{w}\mathbf{T}_{f}$ can be obtained from a pose $\textbf{x}_f$ as $^{w}\mathbf{T}_{f}$~$=$~$\Psi(\textbf{x}_f)$, where $\Psi$ denotes a function that converts a pose into a pose transform matrix. 

Next, a relative pose transform matrix $\Delta \mathbf{T}$ between $\mathcal{V}_{k-M}$ and $\mathcal{V}_{k-M+1}=\mathcal{V}_{f}$ is generated and outliers of matching pairs are rejected at the same time. Suppose that the mean and covariance of $\mathcal{V}_{k-M}$ are $\bar{\mathbf{T}}_{k-M}$ and $\bar{\mathbf{\Sigma}}_{k-M}$, respectively. By using the propagation method, the mean and covariance of $\mathcal{V}_{f}$ are calculated as follows:


\begin{equation}
\begin{aligned}
  ^{w}\mathbf{T}_{f} \leftarrow \bar{\mathbf{T}}_{k-M+1} = \bar{\mathbf{T}}_{k-M}  \Delta \mathbf{T},
  \end{aligned}
  \label{eqn:mean_of_posepropagation_in_GM}
\end{equation}
\vspace{-0.5cm}
\begin{equation}
\begin{aligned}
\mathbf{\Sigma}_{f} \leftarrow \bar{\mathbf{\Sigma}}_{k-M+1} = \mathbf{\bar{\Sigma}}_{0,k-M+1} + \mathbf{\bar{\Sigma}}_{k-M,k-M+1},
 \end{aligned}
\label{eqn:cov_of_posepropagation_in_GM}
\end{equation}
where $\mathbf{\bar{\Sigma}}_{i,j}$ denotes the covariance between the $i$-th and $j$-th nodes. In particular, $\mathbf{\bar{\Sigma}}_{1}$ is $\mathbf{0}_{4\cross 4}$. We set the temporal node $\mathcal{V}_{f}$ using~\textcolor{qwr}(\ref{eqn:mean_of_posepropagation_in_GM}) and~\textcolor{qwr}(\ref{eqn:cov_of_posepropagation_in_GM}), and then a temporal pose graph $\mathcal{G}^{\prime}=(\mathcal{V}^{\prime},\mathcal{E}^{\prime})$ can be generated with $\mathcal{V}^{\prime}=\mathcal{V}_{\text{opt}}~\cup~{\mathcal{V}_{f}}$ and $\mathcal{E}^{\prime}\subset \mathcal{V}^{\prime} \cross \mathcal{V}^{\prime}$. 
Next, the temporal pose graph $\mathcal{G}^{\prime}$ is incrementally pruned and optimized using an efficient online pruning algorithm. 

\subsubsection{Node Pruning and Optimization}

\begin{figure}[t!]
 \vspace{0.25cm}
    \captionsetup{font=footnotesize}
    \centering
	\begin{subfigure}[b]{0.5\textwidth}
		\includegraphics[width=1\textwidth]{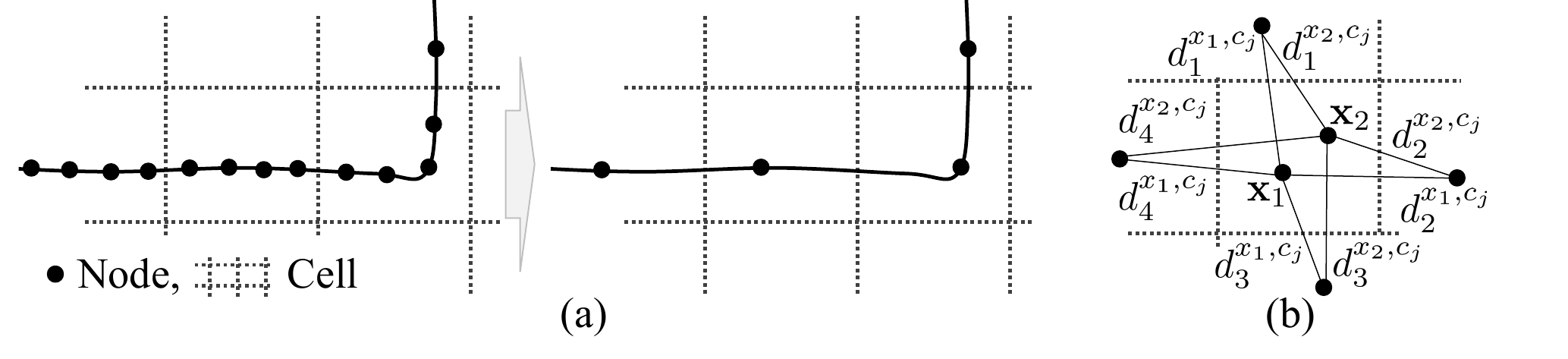}
	\end{subfigure}
	\vspace{-0.7cm}
	\caption{Example of node elimination and weights. (a) Node displacement before and after redundant node elimination; (b) Geometric weights for $\mathbf{x}_1$ and $\mathbf{x}_2$ in cell $c_j$ (e.g. the geometric weight of $\mathbf{x}_{1}$ is $\sum d_{i}^{x_{1},c_j}$ $(i=1,\cdots,4)$).}
	\label{fig:fig_eliminationNodes}
	\vspace{-0.6cm}
\end{figure}

Aiming to maintain a single node in a cell, optimize memory efficiency, and expedite pose optimization, the algorithm selects the most informative node, as illustrated in Fig.~\ref{fig:fig_eliminationNodes}(a). By leveraging uncertainty representation~\cite{rodriguez2018importance} and using a geometrical weight as shown in Fig.~\ref{fig:fig_eliminationNodes}(b), we define the total weight $w_{(c_{j},l)}$ as
\vspace{-0.2cm}
\begin{equation}
\begin{aligned}
 w_{(c_{j},l)} = s \operatorname{Tr}\left( \mathbf{\Lambda}_{(c_{j},l)} \right)+(1-s) \sum_{i=1}^{n} (d_{i}^{x_{l},c_{j}})^{2},
  \end{aligned}
  \label{eqn:weight_of_Node}
  \vspace{-0.25cm}
\end{equation}
where $\mathbf{\Lambda}$, $s$, $n$, and $d_{i}^{x_{l},c_{j}}$ denote an information matrix, a scale factor, the number of neighbor nodes, and distance between the $l$-th node of the $j$-th cell and the $i$-th nodes in adjacent cells, respectively. The subscript $(\cdot)_{(c_{j},l)}$ denotes the $l$-th node in the $i$-th cell. The first and second terms in~\textcolor{qwr}(\ref{eqn:weight_of_Node}) represent the information and geometric weights, respectively. This enables appropriate node distributions for commercial service robots with different purposes: guide robots that prioritize information-based navigation using points of interest (PoI), and cleaning robots that require geometrical coverage through zigzag navigation. Furthermore, to ensure real-time operability, we adopt a method that utilizes information on edges instead of nodes, as in~\cite{kretzschmar2012information,chow1968approximating}. Accordingly, we modified~\textcolor{qwr}(\ref{eqn:weight_of_Node}) as follows:
\vspace{-0.2cm}
\begin{equation}
\begin{aligned}
 w_{(c_{j},l)} = s \sum_{k=1}^{m} \left( \Lambda_{(e_{k},c_{j},l)} \right)+(1-s) \sum_{i=1}^{n} (d_{i}^{x_{l},c_{j}})^{2},
  \end{aligned}
  \label{eqn:re_weight_of_Node}
  \vspace{-0.25cm}
\end{equation}
where $\Lambda_{(e_{k},c_{j},l)}$ and $m$ denote the information of the $k$-th edge connected to the $l$-th node in the $j$-th cell and the number of edges, respectively. 

Given the cell size of a grid map, the highest weighted pose $\mathbf{x}_{l}$ in the cell is selected by $\mathbf{x}_{l}=\argmax_{\mathbf{x}_l} w_{(c_{j},l)}$ and others are eliminated. Covariances of selected nodes are updated by compound approximate transformations~\cite{smith1986representation}. Accordingly, new edges are generated among selected nodes. At this stage, the number of generated edges inevitably increases up to $(\sum_{k=1}^{_{{N}_{\text{edge}}-1}}k)-{{N}_{\text{edge}}}$, where ${N}_{\text{edge}}$ denotes the number of edges connected to the eliminated node. To prune newly added edges, we employ data compression methods~\cite{kretzschmar2012information, carlevaris2014generic}, and Chow-Liu trees~\cite{chow1968approximating} with mutual information~\textcolor{qwr}(\ref{eqn:mutual_information}). In addition, we evaluated several minimum spanning tree (MST) methods such as Chazelle~\cite{chazelle2000minimum,karger1995randomized}, and Kruskal and Prime~\cite{graham1985history} to determine a set of edges with a minimum joint probability distribution.    
Empirically, in our study, an average elimination of 4 nodes and 6 edges per node was observed. Based on these metrics and the computational complexity expressed as $O(N_{\text{edge}}\log (N_{\text{node}}))$~\cite{chazelle2000minimum,karger1995randomized,graham1985history}, where ${N}_{\text{node}}$ is the number of nodes, we chose the Kruskal method, which is known to be suitable for handling simple and sparse trees.

At the optimization stage, we leverage the max-mixture (MM) approach~\cite{olson2013inference} with the pruned graph defined as $\mathcal{G}^{\prime\prime}$~=~$(\mathcal{V}^{\prime\prime},\mathcal{E}^{\prime\prime})$ for a robust SLAM back-end. 
Finally, by applying $\mathcal{V}^{\prime\prime}$ and $\mathcal{E}^{\prime\prime}$ to the nonlinear least-squares problem, our cost function is defined as
\begin{equation}
\vspace{-0.15cm}
\begin{aligned}
  \mathbf{\mathcal{X}}^*=\argmin_{\mathbf{\mathcal{X}} \in \mathcal{V}^{\prime\prime}} \Big{\{} \sum_{i}  \|{h}(\mathbf{x}_{s,i-1},\mathbf{x}_{s,i})-\mathbf{z}_{i-1,i}\|^{2}_{\mathbf{\Sigma}_{s,i}} +\\\sum_{(j,k) \in \mathcal{E}^{\prime\prime} } {\rho} \big{(}\|{g}(\mathbf{x}_{s,j},\mathbf{x}_{s,k})-\boldsymbol{\omega}_{j,k} \|^{2}_{\mathbf{\Omega}_{s,j,k}} \big{)}\Big{\}},
  \end{aligned}
  \label{eqn:costfunc}
\end{equation}
where ${h}(\cdot)$ and ${g}(\cdot)$ are odometry predictions and loop constraints, respectively; $\mathbf{z}_{i-1,i}$ and $\boldsymbol{\omega}_{j,k}$ represent observations pertaining to odometry and loop closure, respectively; $\mathbf{\Sigma}_{s,i}$ and $\mathbf{\Omega}_{s,j,k}$ are combined covariance matrices~\cite{smith1986representation}. 
Our multi-stage approach, named as CLOi-Mapper, has been implemented in several embedded systems and deployed in the commercial products.


\vspace{-0.5cm}
\section{Experiments}\label{sec:Experiments}
\subsection{Experimental Setups and Environments}

We conducted experiments in diverse environments, spanning in-home settings, expansive indoor spaces, offices characterized by repetitive layouts, and varied in-house scenarios across Korea, Germany, and the USA. Our primary objectives included refining pose tracking to ensure local consistency, optimized mapping for global consistency, and systematically testing the applicability of our methodology in real-world settings.

First, we studied an airport guide robot named AirStar\footnote[2]{\url{https://www.youtube.com/watch?v=ztdARyV-Njg}} at Incheon International Airport, Republic of Korea. The study encompassed areas such as airside (A/S), landside (L/S), and baggage claim (B/C) zones within Terminals 1 and 2. \textit{AirStar} is equipped with two monochromatic cameras with $704\cross478$ resolution and a 2D LiDAR sensor that provides measurements ranging from 30 cm to 30 m with 1.0 deg angle resolution at a rate of 1Hz.

Second, we developed an office and home cleaning robot equipped with a single camera ($1280\cross960$ resolution) and an embedded processor (Cortex-A9\footnote[3]{\url{https://developer.arm.com/Processors/Cortex-A9}}). In particular, the office cleaning robot is equipped with a 2D LiDAR sensor with a range of 10 m and 1.0 deg angle resolution.

\begingroup
\begin{table}[t!]
	\centering
         \captionsetup{font=footnotesize}
         \captionsetup{justification=centering, labelsep=newline}       
          \caption{\sc{Quantitative performance comparison of locally consistent trajectories. The sensor combinations \texttt{C}, \texttt{2C}, \texttt{L}, and \texttt{C+L} denote a single camera, two cameras, one LiDAR, and one camera + LiDAR, respectively.}}
    \vspace{-0.1cm}
	\setlength{\tabcolsep}{5pt}
	\begin{tabular}{cccccccc}
    \toprule \midrule
	Place& \# of&Travelled	& \multicolumn{4}{c}{Average pose error ($t_{\text{rel}}$, m)} \\ \cline{4-7}
      (Incheon, Korea)& nodes&distance (m)	    & \texttt{C} &\texttt{2C} &\texttt{L} &\texttt{C+L} \\	\hline
   Terminal 1 L/S & 10,732 & 6,608 & 0.48 & 0.48 & 0.07 & \textbf{0.06}\\
    Terminal 1 A/S & 11,938 & 3,449 & 1.09 & 1.28 & \textbf{0.1} & \textbf{0.1}\\
    Terminal 1 B/C & 9,992 &2,701 & 0.57 & 0.56 & 0.11 & \textbf{0.09}\\  
    Terminal 2 B/C & 4,275 &1,306 & 1.14 & 1.05 & \textbf{0.28} & \textbf{0.28}\\      
    Terminal 2 A/S & 8,913  & 6,335 & 83.2 & 1.62 & 0.78 & \textbf{0.22}\\    
	\midrule\bottomrule
	\end{tabular}
	\label{table:Localization_result}
	\vspace{-0.15cm}
   \vspace{-0.5cm}
\end{table}
\endgroup

\vspace{-0.3cm}
\subsection{Parameters and Hardware Settings}
Empirically, we set $s= 0.5$ in~(\ref{eqn:re_weight_of_Node}) and a submap (see Fig.~\ref{fig:fig_GPT_graph}) size of $4.1 \cross 4.1~\text{m}$ for in-home environments, while we set a submap size of $10 \cross 10~\text{m}$ for large-scale environments. The cameras were tilted up at an angle of $45^\circ$ for the airport guidance robot and $60^\circ$ for the other robots.

\vspace{-0.3cm}
\subsection{Error Metrics}
\vspace{-0.1cm}
As quantitative metrics, average relative pose errors ($t_{\text{rel}}$) are defined as
\begin{itemize}
    \item $t_{\text{rel}}= \sqrt{\sum_{n=1}^{N} (\mathbf{t}_{n, \text{GT}}-{\hat{\mathbf{t}}}_{n})^{2} / N}$,\\    
\end{itemize} \vspace{-0.3cm}
\noindent where $\mathbf{t}_{n,\text{GT}}$ and $\hat{\mathbf{t}}_{n}$ denote the ground truth position vector and estimated position vector, respectively; the subscripts $o$ and $_{\text{GT}}$ denote the origin pose and the value from the ground truth, respectively; $N$ represents the number of selected nodes. In addition, to quantify the degree of change from the original map, we propose a metric called the Average Ratio of Pose Shift (ARPS), defined as

\begin{itemize}
    \item $\text{ARPS}=\{(\sum_{i=1}^{M}\frac{|\mathbf{t}_{i,\text{ori}}-\mathbf{t}_{i,\text{pru}}|}{|\mathbf{t}_{i,\text{ori}}|})/M\}\cross 100$,        
\end{itemize} 
\noindent where $\mathbf{t}_{i,\text{ori}}$ denotes the original position vector of the $i$-th node in the map, and $\mathbf{t}_{i,\text{pru}}$ denotes the position vector of the same node after the pruning and optimization stage.

\vspace{-0.3cm}
\section{Experimental Results and Discussion}\label{sec:Experimental Results}
We show the performance evaluation results in terms of our extensible framework, locally consistent poses, and globally consistent poses.
\vspace{-0.3cm}
\subsection{Performances Relative to Combinations of Sensors}

\begin{figure}[b!]
\vspace{-0.5cm}
    \captionsetup{font=footnotesize}
    \centering
	\begin{subfigure}[b]{0.48\textwidth}
		\includegraphics[width=1\textwidth]{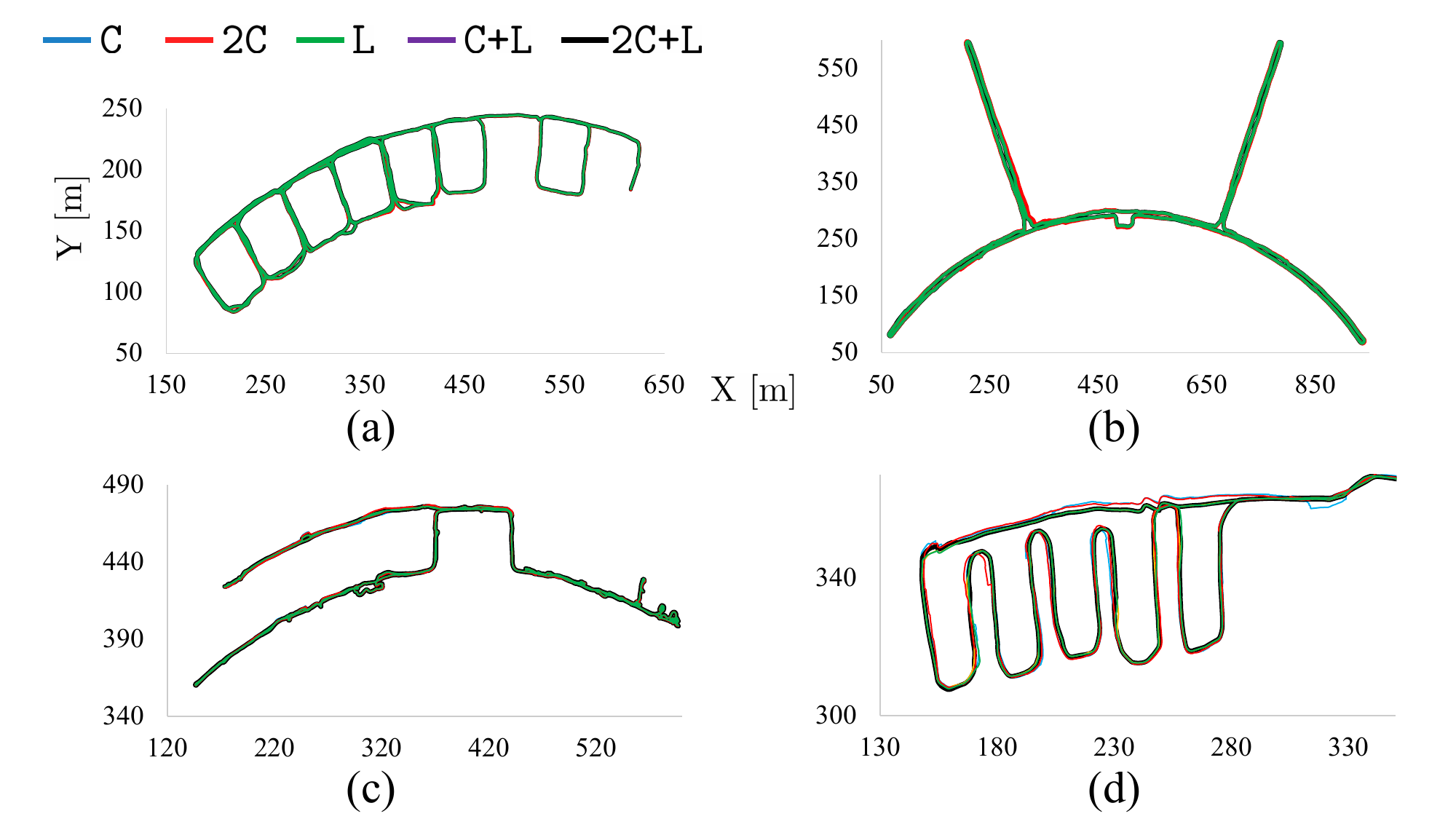}
	\end{subfigure}
	\vspace{-0.1cm}
	\caption{Qualitative comparison of locally consistent trajectories for various sensor combinations at various positions: (a) Terminal 1 L/S, (b) Terminal 1 A/S (c) Terminal 1 B/C, (d) Terminal 2 B/C.}
	\label{fig:fig_Casestudy__2__OfGuidanceRobot}
	\vspace{-0.2cm}
\end{figure}

\begin{figure}[b!]
\vspace{-0.3cm}
    \captionsetup{font=footnotesize}
    \centering
	\begin{subfigure}[b]{0.48\textwidth}
		\includegraphics[width=1\textwidth]{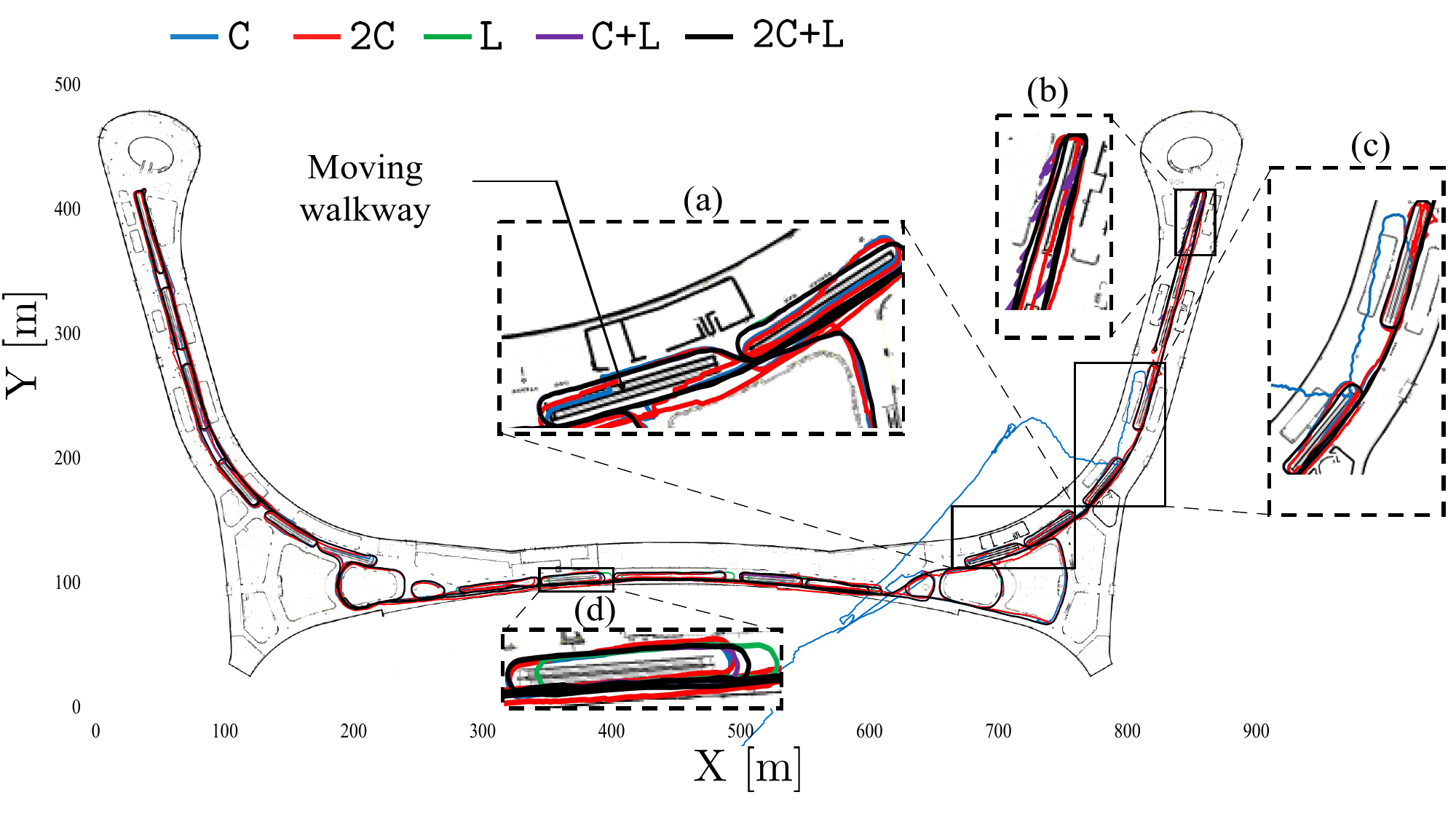}
	\end{subfigure}
	\vspace{-0.3cm}
	\caption{Qualitative comparison of locally consistent trajectories for various sensor combinations at Terminal 2 A/S are presented in Table~\ref{table:Localization_result}. Our algorithm demonstrates consistent pose estimation across multiple case studies involving limited sensor combinations. The case studies are as follows: (a) The trajectory of the 2-camera setup (red line) crosses the moving walkway. (b) Local pose oscillations occur even in the 1 camera + 1 LiDAR scenario (purple line). (c) The trajectory from a single camera's data yields inaccurate results. (d) The trajectory of the LiDAR sensor (green line) intersects the moving walkway. }
	\label{fig:fig_CasestudyOfGuidanceRobot}
\end{figure}

\begin{figure*}[t!]
    \vspace{0.25cm}
    \captionsetup{font=footnotesize}
    \centering
	\begin{subfigure}[b]{0.8\textwidth}
		\includegraphics[width=1\textwidth]{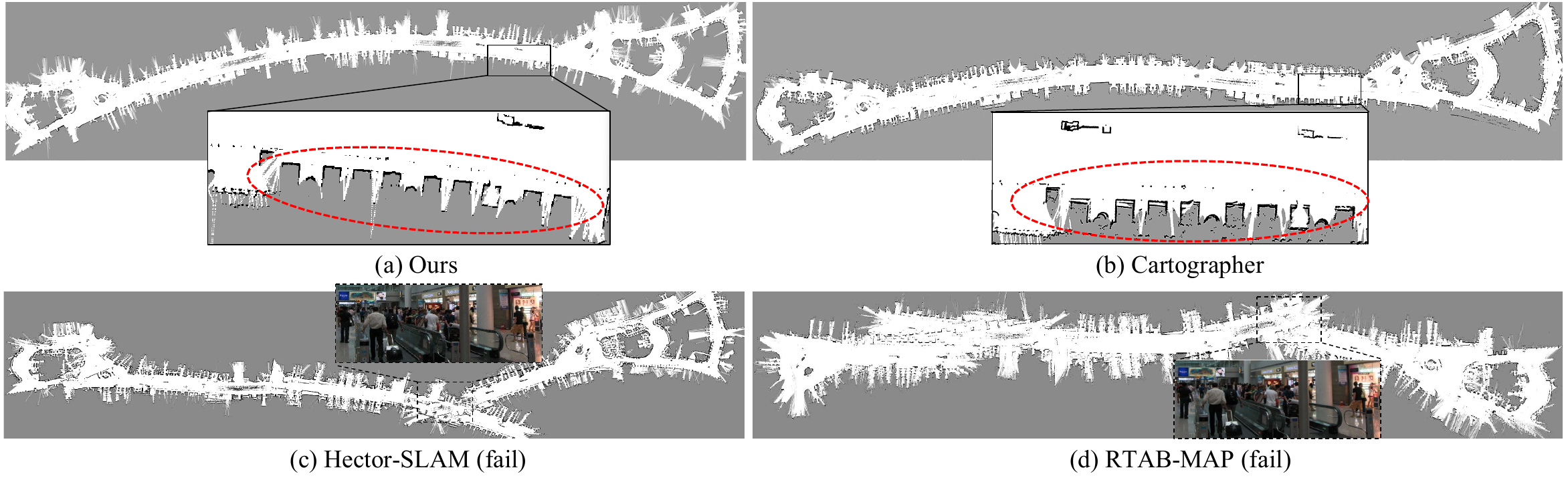}
	\end{subfigure}
	\vspace{-0.2cm}
	\caption{Qualitative comparison of the global map generated by several SLAM algorithms in a section of Terminal 2 A/S, Incheon International Airport. (a) Our result demonstrates consistent performance. (b) Cartographer subtly underperformed compared with our result, which is highlighted by red dotted ellipses indicating wall alignment performance. (c)--(d) Hector-SLAM and RTAB-MAP failed to map after passing the moving walkway. Furthermore, ORB-SLAM3 encountered failures in our case, potentially due to limitations in the number of features ($\leq$ 20) and frequent directional changes.}
  	\label{fig:fig_ComparisonResultsOfGuidanceRobot}
	\vspace{-0.5cm}
\end{figure*}

First, performance changes in terms of locally consistent pose estimation were analyzed. This letter presents the performance of locally consistent pose estimation for various combinations of sensors across five locations, as presented in Table~\ref{table:Localization_result}. Here, the robot has multiple sensors for localization, such as two cameras and a LiDAR sensor. The combination of sensors could be altered owing to various causes, such as a contaminated lens causing a blurry image. Our approach has been validated in complex and expansive environments to address these variations, as illustrated in Figs.~\ref{fig:fig_Casestudy__2__OfGuidanceRobot}~and~\ref{fig:fig_CasestudyOfGuidanceRobot}. Meanwhile, obtaining ground-truth poses is challenging, as it requires high-quality and high-cost equipment. In order to assess the consistency of our algorithm, we conducted trajectory comparisons between the estimations of limited sensor combinations and the trajectory determined using all sensors on our robot (depicted as the black line in Fig.~\ref{fig:fig_CasestudyOfGuidanceRobot}).

\begingroup
\begin{table}[t!]
	\centering
	\captionsetup{font=footnotesize}
	\captionsetup{justification=centering, labelsep=newline}                 
	\caption{\sc{Comparison of performance with widely used SLAM methods suitable for each commercially affordable sensor.}}
	\vspace{-0.1cm}
	\setlength{\tabcolsep}{5pt}
	\begin{tabular}{cccccc}
	\toprule \midrule
	\multicolumn{2}{c}{(@Terminal 2 A/S)} & \multicolumn{4}{c}{Average pose error ($t_{\text{rel}}$, m)}                                          \\ \cline{3-6}
	Category& Algorithm &  \texttt{C} &\texttt{2C} &\texttt{L} &\texttt{C+L}                   \\
	\midrule \midrule
	&  Ours&  $\triangle$ & 1.62 & 0.78 &  0.22        \\
	\midrule
	Vision &ORB-SLAM& \ding{53}  &   \ding{53}   & \multicolumn{2}{c}{\cellcolor{gray!20}}  \\
	\midrule
	\multirow{2}{*}{LiDAR} &  Cartographer & \multicolumn{2}{l}{\multirow{2}{*}{\cellcolor{gray!20}}} & 0.83 & \multirow{2}{*}{\cellcolor{gray!20}}  \\ 
	&Hector-SLAM& \multicolumn{2}{l}{\cellcolor{gray!20}}                  & $\triangle$ & \cellcolor{gray!20}   \\
	
	\midrule 
	Fusion &  RTAB-MAP &\cellcolor{gray!20} &\ding{53} &$\triangle$&$\triangle$                   
	\\ \midrule\bottomrule
	\vspace{-0.2cm}
	\end{tabular}

	\raggedright $\triangle$ and \ding{53} denote map generation with pose errors exceeding 2 meters and map generation failed, respectively.
	\vspace{-0.6cm}
	\label{table:ComparisonofLocalization_result}
\end{table}
\endgroup
	
Specifically, the result obtained from a single camera at Terminal 2 A/S (83.2 m) is inconsistent with the others, as depicted by the blue lines in Fig.~\ref{fig:fig_CasestudyOfGuidanceRobot}. This is caused by a significant amount of wrongly associated poses, which could be due to similarity in captured images in repeatedly patterned environments. Consequently, the proposed algorithm effectively achieved pose consistency by seamlessly integrating multiple sensors within a large-scale environment.
Furthermore, the proposed algorithm's generalizability in service robots was comprehensively evaluated across diverse in-home environments in three countries, namely, Korea, the USA, and Germany, with home cleaning robots. Accordingly, the proposed algorithm was verified to estimate a current pose in real-time. The performance of local consistency related to cleaning robots in in-home environments has been proven in real-world cases (with approximately 500,000+ units sold) by users. Therefore, we believe that the evaluation is sufficient. This method was proven highly valuable for coverage navigation, particularly in scenarios that require precise path following, such as navigating a zigzag path, even during the mapping stage.


\vspace{-0.5cm}
\subsection{Global Consistency Compared With Widely Used Methods Suitable for Commercially Affordable Sensors}

Regarding global consistency, compared with widely used SLAM algorithms in the industry fields, the proposed algorithm can be more effective and applicable within crowded and large-scale environments, as illustrated in Fig.~\ref{fig:fig_ComparisonResultsOfGuidanceRobot}. Despite numerous attempts, even the widely used algorithms face challenges in completing the entire map, often resulting in failure,  as presented in Table~\ref{table:ComparisonofLocalization_result}. The results for each sensor using the proposed method were compared separately to highlight its effectiveness. This issue appears to emerge from the algorithm's inability to satisfy the requirements related to parameter settings (e.g. lack of features). 

\begin{figure}[t!]
\centering   
    \captionsetup{font=footnotesize}

	\begin{subfigure}[t]{0.16\textwidth}
		\includegraphics[width=1\textwidth]{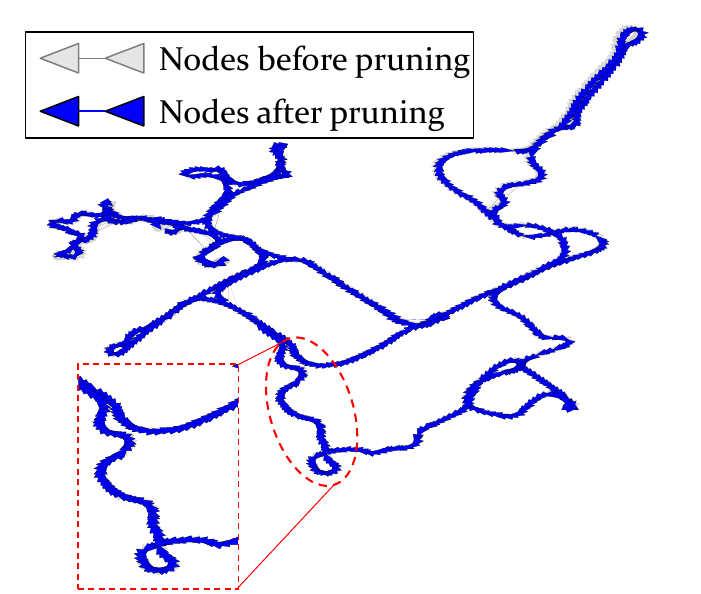}
            \caption{}           
	\end{subfigure}~ 
        \begin{subfigure}[t]{0.16\textwidth}
        \centering
		\includegraphics[width=1\textwidth]{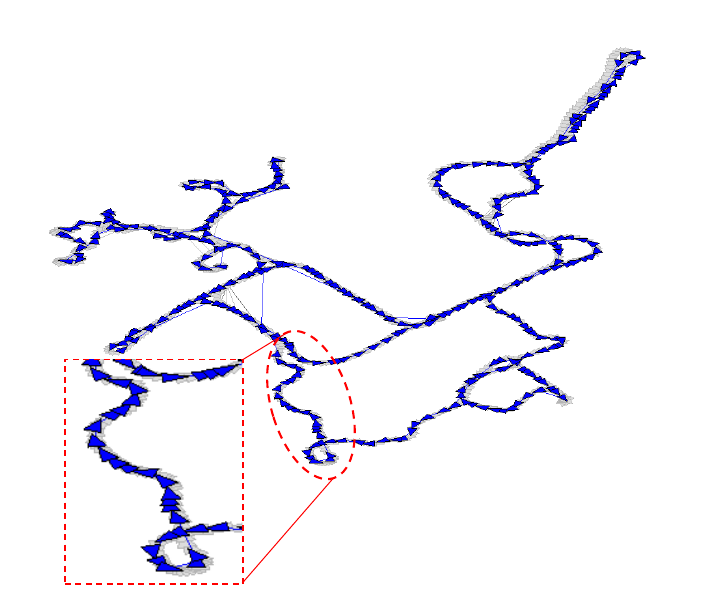}
            \caption{}           
	\end{subfigure}~
  	\begin{subfigure}[t]{0.16\textwidth}
        \centering  
	   \includegraphics[width=1\textwidth]{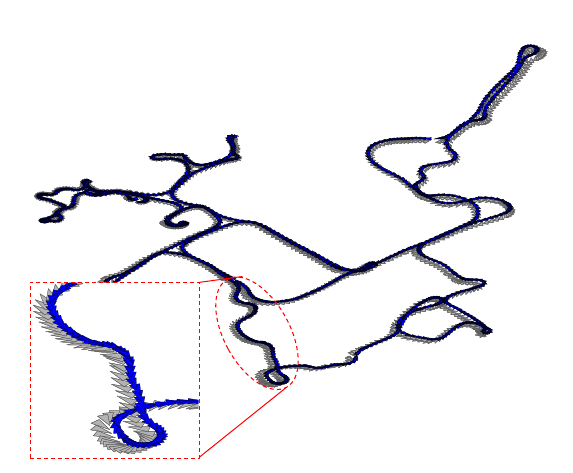}
         \caption{}           
       \end{subfigure}		
 	
\caption{Qualitative results of pose-graph pruning and optimization for CSAIL dataset. (a) Our result with a grid cell size of 0.3$\cross$0.3 m; (b) Our result with a grid cell size of 1$\cross$1 m set by the method described in Fig.~\ref{fig:fig_eliminationNodes}; (c) The result based on the recent data-compression algorithm~\cite{9981584}.}
\label{fig:Resultofpruning}
\vspace{-0.25cm}
\end{figure}

\begingroup
\begin{table}[t!]
	\centering
	  \captionsetup{font=footnotesize}
         \captionsetup{justification=centering, labelsep=newline}               
	\caption{\sc{Performance comparison of node pruning with various public datasets and our dataset.}}
	\vspace{-0.15cm}
	\setlength{\tabcolsep}{5pt}
	\begin{tabular}{l|ccccc}
	
	\toprule \midrule
              & \# of $\text{Node}$ & \# of $\text{Edge}$ & NPC & $\text{ARPS}^{4}$\\              
         \midrule        
        CSAIL &  ~~327 (1,045) & ~~354 (1,172) & 1 & 0.78\% \\
        FR079 &  ~~~~718 (989) & ~~861 (1,217) & 1 & 0.60\% \\        
        M3500 &  1,113 (3,500) & 1,762 (5,453) & 1 & 4.20\% \\
        R9-home{$^\mathsection$}  & ~~~~317 (600)  & ~~663 (1,038) & 1 & 1.30\%     
        \\\midrule\bottomrule
	\end{tabular}
	\label{table:pruning_result}
	\vspace{-0.1cm}
 \begin{flushleft}    
	Note: The numbers in parentheses are the results before pruning. NPC denotes the average numbers of nodes per grid cell. $\mathsection$ represents a dataset with our cleaning robot called R9 (https://github.com/Multiplanet-Robot).
    \end{flushleft}
	\vspace{-0.8cm}
\end{table}
\endgroup

Furthermore, we qualitatively compared our SLAM back-end with the recent pruning algorithm~\cite{9981584}, as illustrated in Fig.~\ref{fig:Resultofpruning}. The proposed pruning algorithm was evaluated on various datasets as represented in Table~\ref{table:pruning_result}, and verified that the results were aligned with our objectives in terms of commercialization; specifically, the criteria were satisfied: the number of nodes per cell $\leq$ 2, number of edges connected to a node $\leq$ 3, and ARPS after pruning and optimization was less than 10 \%. We found that the method using~\textcolor{qwr}(\ref{eqn:re_weight_of_Node}) significantly enhanced the computation time on public datasets rather than using~\textcolor{qwr}(\ref{eqn:weight_of_Node}), such as the Manhattan (M3500; 3,551 ms $\rightarrow$ 10.36 ms) and CSAIL datasets (195.67 ms $\rightarrow$ 2.79 ms). Moreover, the average ratio of pose shift after optimization was similar to the results obtained using~\textcolor{qwr}(\ref{eqn:weight_of_Node}) (M3500: 3.62~\% $\rightarrow$ 4.20~\%, CSAIL: 3.3~\% $\rightarrow$ 0.78~\%), with smaller values indicating better performance.

\begin{figure}[t!]
    \captionsetup{font=footnotesize}
    \centering
	\begin{subfigure}[b]{0.45\textwidth}
		\includegraphics[width=1\textwidth]{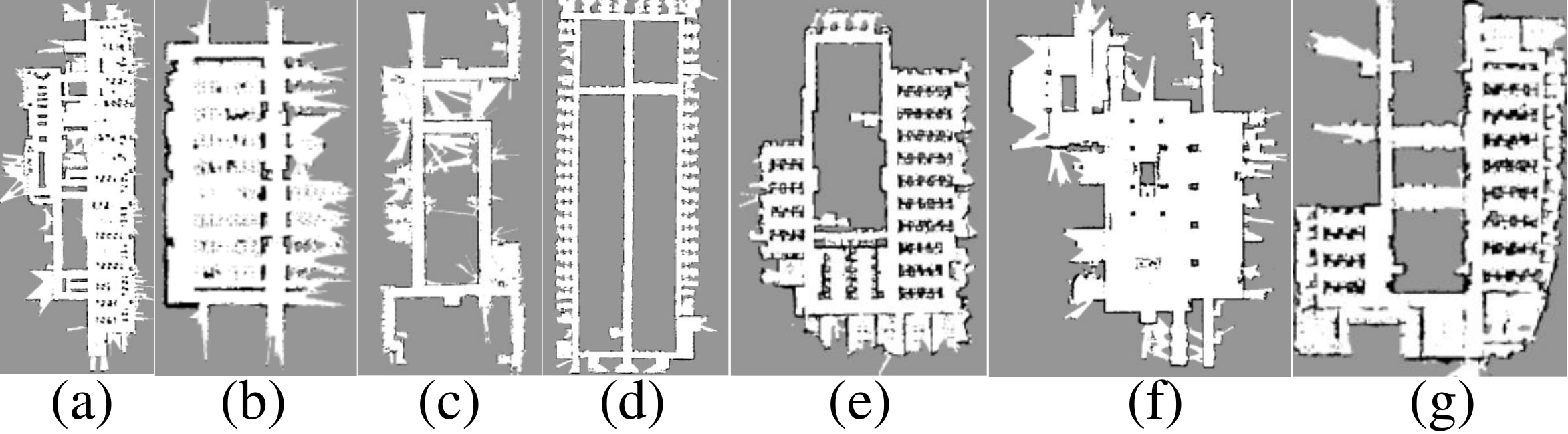}
	\end{subfigure}
        \vspace{-0.1cm}
	\caption{Illustration of qualitative mapping results corresponding to large-scale office environments in Korea with repeated patterns such as partitions; (a) Nexen building 4F, (b) LG Science Park 7F, (c) LGE Seocho 14F, (d) Science Park 6F, (e) SNI building 3F, (f) LGE Seocho 1B, and (g) Seoul building 13F.}
	\label{fig:fig_Qualitative_ResultsOfofficeEnv}
	\vspace{-0.5cm}
\end{figure}

In addition, the proposed algorithm was evaluated in terms of the consistency of global maps by applying it to a robot (B9) in various real office environments. This letter presents only the qualitative results, as illustrated in Fig.~\ref{fig:fig_Qualitative_ResultsOfofficeEnv}. Consequently, the applicability of the proposed algorithm in offices has been verified successfully.

\vspace{-0.5cm}
\subsection{Resource Usage in the Embedded System}
Furthermore, in this study, our SLAM algorithm is operated on an embedded processor called Cortex-A9, with processor usage below 25\%. In addition, memory usage is less than 180 MB to cover a 330 $\text{m}^2$ space. For example, the GPE is operated at over 3 Hz on the specified processor. 
\vspace{-0.2cm}
\section{Conclusions} \label{sec:conclusion}
This study has presented a multi-stage approach to global pose estimation and mapping, particularly suitable for low-cost embedded systems. The proposed method seamlessly integrates various sensors into our SLAM algorithm without requiring significant structural modifications. We proposed a memory-efficient back-end with pruning and optimization techniques, which were validated across multiple mass-produced commercial robots in terms of consistency and stability. Future research will extend this framework to multi-robot SLAM.

\bibliographystyle{IEEEtran}
\vspace{-0.5cm}
\bibliography{./ral24,./IEEEabrv}

\end{document}